\documentclass{article} 
\usepackage{iclr2025_conference,times}

\input{math_commands.tex}

\usepackage{graphicx}
\usepackage{multirow}
\usepackage{booktabs}
\usepackage[pagebackref=true,breaklinks,colorlinks,citecolor=blue,linkcolor=blue,urlcolor=blue,hypertexnames=false]{hyperref}

\title{A Simple Approach to Unifying Diffusion-based Conditional Generation}

\author{Xirui Li$^{1}$ \qquad Charles Herrmann$^{2}$ \qquad Kelvin C.K. Chan$^{2}$ \qquad Yinxiao Li$^{2}$ \\
\textbf{Deqing Sun}$^{2}$ \qquad \textbf{Chao Ma}$^{1}$\thanks{Corresponding author.} \qquad \textbf{Ming-Hsuan Yang}$^{2}$ \\
$^{1}$
Shanghai Jiao Tong University \qquad  $^{2}$ Google DeepMind\\
{\small Project webpage: \url{https://lixirui142.github.io/unicon-diffusion/}}
}

\iclrfinalcopy 
\begin{document}

\maketitle
\vspace*{-8mm}
\begin{center}
\includegraphics[width=1.0\linewidth]{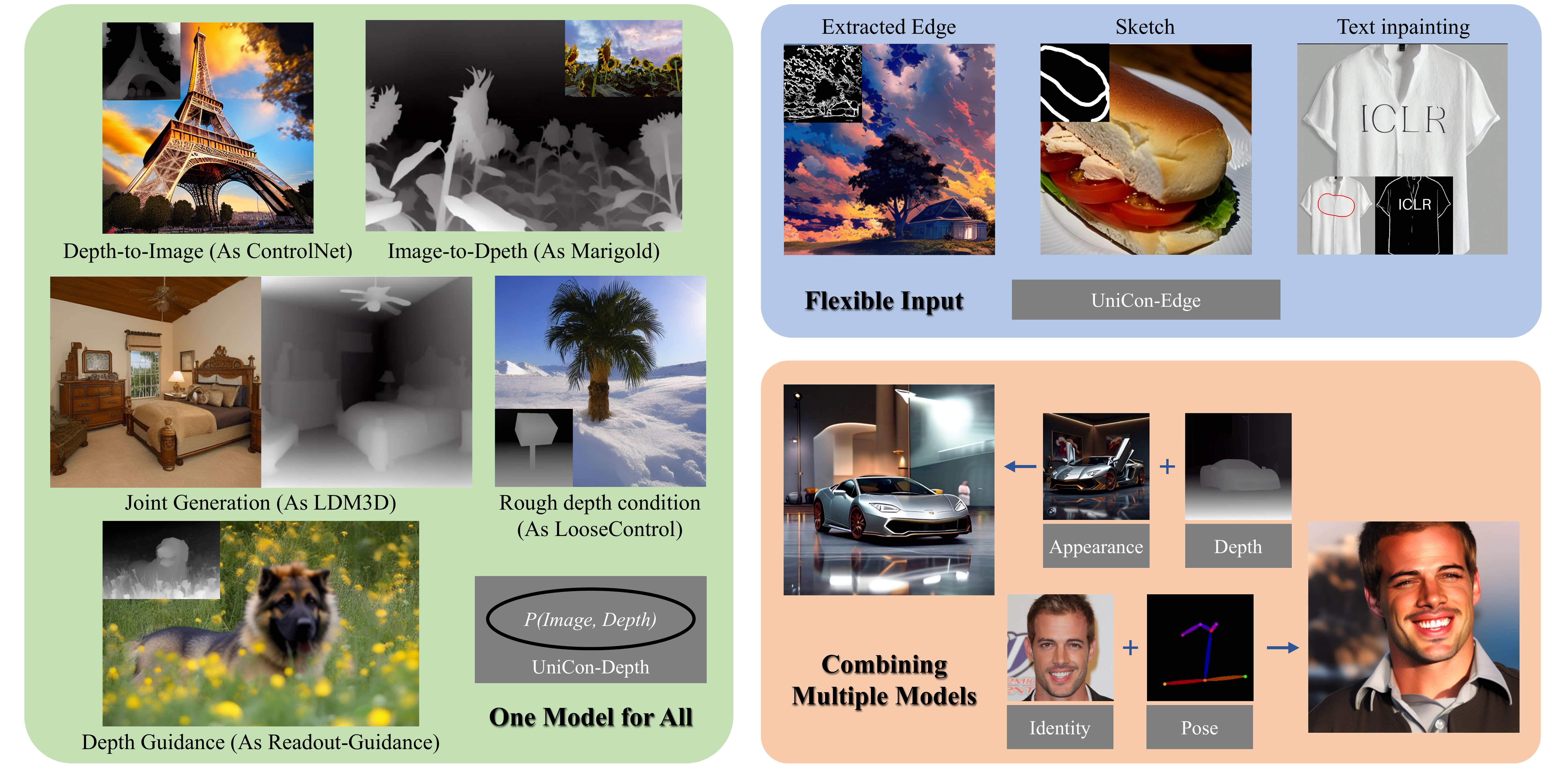}
\captionof{figure}{The proposed \ourmethod{} supports diverse generation behavior in one model for a targeted type of image and condition.  \ourmethod{}  also offers flexible conditional generation ability with natural support for free-form input and seamless integration of multiple models.}
\label{fig:teaser}
\end{center}

\begin{abstract}

Recent progress in image generation has sparked research into controlling these models through condition signals, with various methods addressing specific challenges in conditional generation. Instead of proposing another specialized technique, we introduce a simple, unified framework to handle diverse conditional generation tasks involving a specific image-condition correlation. By learning a joint distribution over a correlated image pair (\eg image and depth) with a diffusion model, our approach enables versatile capabilities via different inference-time sampling schemes, including controllable image generation (\eg depth to image), estimation (\eg image to depth), signal guidance, joint generation (image \& depth), and coarse control. Previous attempts at unification often introduce significant complexity through multi-stage training, architectural modification,
or increased parameter counts. In contrast, our simple formulation requires a single, computationally efficient training stage, maintains the standard model input, and adds minimal learned parameters (15\% of the base model). Moreover, our model supports additional capabilities like non-spatially aligned and coarse conditioning. Extensive results show that our single model can produce comparable results with specialized methods and better results than prior unified methods. 
We also demonstrate that multiple models can be effectively combined for multi-signal conditional generation.

\end{abstract}

\section{Introduction}
\label{sec:intro}

Text-to-image diffusion models, such as Dall-E 2~\citep{DALLE2}  and Imagen~\citep{ho2022imagen}, have revolutionized the field of image generation, leading to contemporary models~\citep{midjourney, baldridge2024imagen} that can produce images almost indistinguishable from real ones.
This progress in generative modeling, particularly with diffusion models, has spawned new research areas and reshaped existing fields within computer vision.
With advancements in image quality, the generative community has expanded its focus to controllability, resulting in many different approaches, each promoting a distinct scheme for guiding the generative process.
ControlNet~\citep{zhang2023adding} highlights the effectiveness of using modalities like depth and edges as conditional input. Meanwhile, other works, such as Loose Control~\citep{bhat2024loosecontrol} and Readout Guidance~\citep{luo2024readout}, propose alternative conditioning types (\eg coarse depth maps) and control mechanisms (\eg guidance through a prediction head).
Concurrently, the estimation community has seen diffusion models advance the state-of-the-art for predicting various modalities from RGB images, \eg Marigold~\citep{ke2023repurposing} repurposes a pretrained image generator to generate depth instead. In addition, other work~\cite {stan2023ldm3d} has explored joint diffusion, generating paired image and depth simultaneously.

Although typically addressed as separate tasks within distinct communities, these problems share a common underlying structure: conditional generation between correlated images. 
Consider the relationship between an image and its depth map: controllable generation translates depth to image, estimation maps image to depth, guidance uses depth predictions to guide image generation,
and joint generation produces image-depth pairs.
This observation motivates us to unify all these tasks under a global distribution modeling problem.
While a few works~\citep{qi2024unigs, zhang2023jointnet} have also explored unified models capable of handling these diverse tasks, they often introduce significant complexity through multi-stage training, increased parameter counts, or architectural modifications. This additional complexity makes creating and using these models difficult, hindering their adoption.

In this paper, we propose \ourmethod{}, a unified diffusion model that learns an image-condition joint distribution with a flexible model architecture and simple but effective training strategy to support diverse inference behaviors.
We propose an architecture adaptation to the standard image generator diffusion model that is more flexible than ControlNet (allowing for non-pixel-aligned conditioning signals) and more efficient to train, decreasing both the number of learned parameters and required training samples.
Inspired by Diffusion Forcing~\citep{chen2024diffusion}, we use a training scheme that disentangles the noise sampling of the image and the condition, allowing flexible sampling strategies at inference time to achieve different conditional generation tasks without explicit mask input.

As shown in Fig.~\ref{fig:teaser}, with the same model but different sampling schedules, \ourmethod{} can do: 1) controllable image generation in the form of ControlNet, Readout Guidance, and Loose Control, 2) estimation, and 3) joint generation.
We train our models based on a large text-to-image model for several different modalities (depth, edges, human poses, identity) and show that the behavior of our single model is similar to or better than specialized methods using standard image quality and alignment metrics. We demonstrate significant improvements over prior unified models in conditional generation, training efficiency, and generation flexibility.
We also show that \ourmethod{}  can combine multiple models for multi-signal conditional generation or switch our model to other base model checkpoints. Our models are trained in about 13 hours on 2-4 Nvidia A800 80G GPUs, adding 15\% parameters to the base model.

The main contributions of this work are:
\begin{compactitem}
\item Proposing a framework that unifies controllable generation, estimation, and joint generation, including model adaptation, training strategy, and sampling methods allowing flexible conditional generation at inference.
\item Demonstrating that our architecture and training can work on a large-scale text-to-image diffusion model with a small number of learned parameters and a relatively small training data scale.
\item Showing that our unified models can perform similar to specialized methods or better than current unified approaches on different modalities.
\end{compactitem}

\section{Related Work}
\label{sec:rel}

\noindent \textbf{Controllable Generation.}
Fine-tuning text-to-image diffusion models to conditional image generation on signals beyond text has gained significant popularity~\citep{huang2023composer, zhang2023adding, ye2023ip, mou2023t2i, sohn2023styledrop}.
ControlNet~\citep{zhang2023adding} trains a control network attached to pre-trained diffusion models to incorporate condition signals, such as edge maps, segmentation masks, and pose estimation.
Based on ControlNet, LooseControl~\citep{bhat2024loosecontrol} generalizes depth conditioning to loose depth maps that specify scene boundaries and object positions.
Readout-Guidance~\citep{luo2024readout} proposes a new control scheme by adding prediction heads to internal features and guiding the generation by the predicted condition signal.
DAG~\citep{kim2022dag} guides a diffusion model to generate geometrically plausible images using depth prior.
Conditional editing tasks, such as DiffEdit ~\citep{couairon2023diffedit}, have enhanced conditional image manipulation by applying diffusion-based models for inpainting and editing tasks. 
Instead of one specific control behavior, our proposed framework provides diverse controlling abilities through different sampling strategies.

\noindent \textbf{Estimation.} Extracting signals like depth, surface normals, or segmentation maps from RGB images has been a longstanding challenge in computer vision. 
Typically, each task has been addressed in isolation or limited combinations,\eg joint depth and segmentation, but distinct from image generation.
Starting works like DDVM~\citep{saxena2023ddvm}, these tasks have started to be addressed directly with diffusion models including depth prediction~\citep{saxena2023zero,saxena2023ddvm}, optical flow prediction~\citep{saxena2023ddvm},
correspondence matching~\citep{Nam:2024:DMD}, \etc. Recently, there has been considerable interest in either using generative features inside estimators~\citep{xu2023open, zhao2023unleashing} or explicitly fine-tuning image generators as estimators, such as  Marigold~\citep{ke2023repurposing} which adds a clean RGB conditioning and fine-tunes the entire model to diffuse a depth map. While Marigold shares some similarities with our depth estimation setting for our RGBD model, our approach differs significantly in both goals and techniques.  Unlike Marigold, which focuses solely on depth and discards image generation capabilities, our method retains the ability to perform image generation, depth estimation, and other tasks, through lightweight LoRA fine-tuning.

\noindent \textbf{Joint and Unified Generation.}
While less common than controllability or estimation, several works have attempted to unify multiple images and modalities within a single model. LDM3D~\citep{stan2023ldm3d} jointly generates image and depth data in an RGBD latent space. Following approaches commonly include an inpainting mask in the input to extend joint generation to bidirectional conditional generation. For example, UniGD~\citep{qi2024unigs} unifies image synthesis and segmentation through a diffusion model trained with image, segmentation, and inpainting mask as inputs. 
JeDi~\citep{zeng2024jedi} learns a joint distribution over images that share a common object, facilitating personalized image generation. Among the most relevant works, JointNet~\citep{zhang2023jointnet} adopts a symmetric ControlNet-like structure for generating both image and depth, utilizing an inpainting scheme to support depth-to-image and image-to-depth generation.
Our approach presents several improvements over these unified methods. First, our training strategy enables flexible conditional generation without requiring an explicit mask input. Thus we can avoid concatenating multiple inputs in the feature dimension or adding inpainting masks, which allows our model to act like adapters that can be plugged into the base model checkpoints. Furthermore, our structure supports both loosely correlated image pairs (as in JeDi) and densely correlated pairs (as in JointNet), providing more versatile capabilities across different scenarios.

\section{Preliminaries}
\label{sec:prelim}
\noindent \textbf{Diffusion Model.}
Diffusion models~\citep{sohl2015deep,ho2020denoising, song2020score} are generative models that model a data distribution $p(\rvx)$ through an iterative denoising process.
Consider a forward process gradually adding Gaussian noise $\epsilon$ to data $\rvx_0 \sim p(\rvx)$ with timesteps $t=1,\dots,T$ and noise schedule $\{\alpha_t\}$,
\begin{equation}
    q(\rvx_t|\rvx_0) = \mathcal{N}(\rvx_{t}; \sqrt{\bar{\alpha}_t} \rvx_0, (1-\bar{\alpha}_t) \mI), \quad q(\rvx_t|\rvx_{t-1}) = \mathcal{N}(\rvx_{t}; \sqrt{\alpha_t} \rvx_{t-1}, (1-\alpha_t) \mI),
    \label{eq:diff_forward}
\end{equation}
where $\bar{\alpha}_t = \prod_{s=1}^{t} \alpha_s$ and $x_T \sim \mathcal{N}(0, \mI)$ reaches pure noise.
Diffusion models learn to denoise $x_t$ at any timestep by estimating $\hat{\rvx}_{\theta}(\rvx_t,t) \approx \rvx_0$. According to common $\bm{\epsilon}$-parameterization, we can train the model to predict the noise $\bm{\epsilon}_{\theta}(\rvx_t,t)$ instead using the following least squares objective,
\begin{equation}
    \min_{\theta} E_{\rvx_0, \bm{\epsilon},t} \| \bm{\epsilon} - \bm{\epsilon}_{\theta} (\rvx_t,t)\|^2,
    \label{eq:diff_obj}
\end{equation}
where $\rvx_t = \text{AddNoise}(\rvx_0, t) = \sqrt{\bar{\alpha}_t} \rvx_0 + \sqrt{1-\bar{\alpha}_t} \bm{\epsilon}$ and $\bm{\epsilon} \sim \mathcal{N}(0, \mI)$ is random noise. With the trained denoiser, one can adopt any sampler~\citep{ho2020denoising, song2020denoising, karras2022elucidating} to sample new data from noise.
%
Recent latent diffusion models~\citep{rombach2022high,  ramesh2022hierarchical} map image data into the latent space to improve performance and efficiency. We base our experiments on Stable Diffusion~\citep{rombach2022high}, a large-scale text-to-image latent diffusion model.


\begin{figure}[t]
\begin{center}
\includegraphics[width=\linewidth]{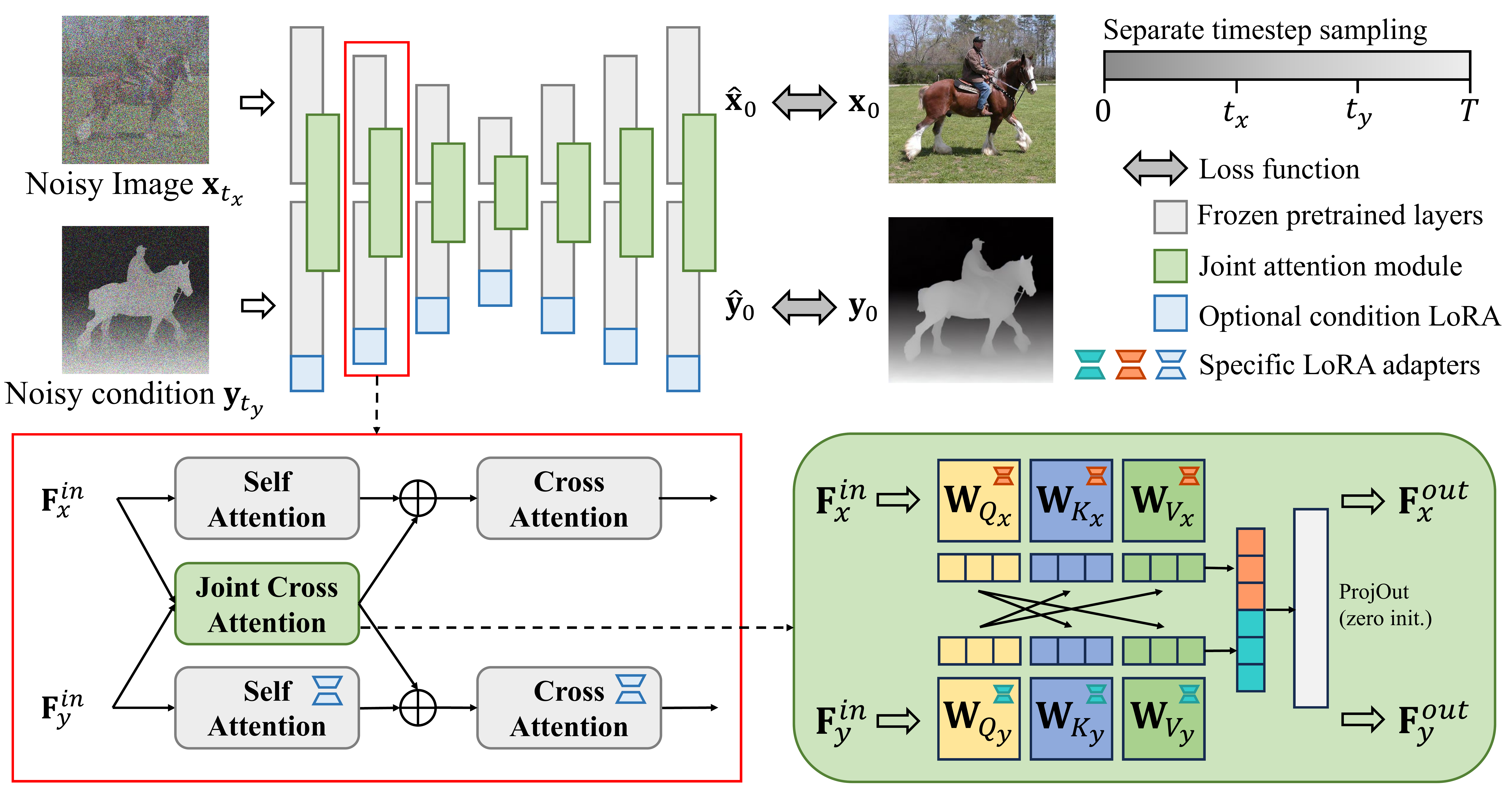}
\end{center}
\caption{\textbf{\ourmethod{}\ pipeline.} Given a pair of image-condition inputs, our \ourmethod{} model processes them concurrently in two parallel branches, with injected joint cross-attention modules where features from two branches attend to each other. We use LoRA weights to adapt our model from a pretrained diffusion model. During training, we separately sample timesteps for each input and compute loss over both branches.
}
\label{fig:pipeline}
\end{figure}

\section{Method}
\label{sec:method}

Our method, \ourmethod{}, aims to train a unified diffusion model for diverse conditional image generation tasks, such as conditional generation on clean, coarse, or partial control signals, estimation, and joint generation.
The key idea is to learn a joint distribution over a correlated image pair $\rvx, \rvy$, which allows flexible conditional sampling.
The image pair can have strict spatial alignment (image-depth, image-edge) or loose semantic correspondence (frames from one video clip).
In Sec.~\ref{sec:motivation}, we first introduce our motivation for learning the joint distribution.
Then, we elaborate on our model structure and training pipeline in Sec.~\ref{sec:model_struct} and the sampling strategies for flexible conditional generation in Sec.~\ref{sec:infer}.

\subsection{Motivation}
\label{sec:motivation}

Image diffusion models offer significant flexibility when sampling in the learned image distribution $p(\rvx)$.
In addition to generating new image $\rvx \sim p(\rvx)$, one can perform image inpainting by conditioning partial image $\rvx^m$ and image editing by conditioning noisy image $\rvx_t$, corresponding to sampling in conditional distributions $p(\rvx|\rvx^{m})$ and $p(\rvx|\rvx_t)$.
Our motivation is to generalize these abilities from a single image to a correlated image pair $(\rvx, \rvy)$ by learning a joint distribution $p(\rvx, \rvy)$. We then use the modeled joint distribution to enable flexible conditional generation.
If the conditional signal is encoded as image $\rvy$, we can train a diffusion model to denoise both image $\rvx$ and condition $\rvy$.
The trained joint diffusion model supports various conditional generation tasks that can be unified as sampling in the following conditional distribution,
\begin{equation}
    (\rvx,\rvy) \sim p(\rvx,\rvy|\rvx_{t_x}^{m_x}, \rvy_{t_y}^{m_y}),
    \label{eq:joint_cond_distribution}
\end{equation}
where $\rvx_{t_x}^{m_x}$ indicates $\rvx$ partially masked by mask $m_x$ under noise level (timestep) $t_x$.

Sampling in the joint conditional distribution can be regarded as a direct generalization of image inpainting and editing.
The conditional generation and estimation are equivalent to inpainting image $\rvx$ or condition $\rvy$  ($p(\rvx|\rvy_0), p(\rvy|\rvx_0)$).
We can also inpaint image and condition for partial control ($p(\rvx,\rvy|\rvx_0^{m_x}, \rvy_0^{m_y})$).
Adding noise to $\rvy$ enables the model to accept coarse condition input ($p(\rvx,\rvy|\rvy_t)$) like SDEdit~\citep{meng2021sdedit} does for image. We can control the condition fidelity by adjusting the noise level.
To sum up, combining spatial masking $m_x, m_y$ and noise masking $t_x, t_y$ provides substantial possibilities in free-form conditional generation.

Based on this motivation, our goal is to train a unified diffusion model for targeted image pair $(\rvx, \rvy)$ and develop sampling strategies to support the conditional sampling described in Eq.~\ref{eq:joint_cond_distribution}.

\subsection{\ourmethod{}}
\label{sec:model_struct}

Figure~\ref{fig:pipeline} illustrates the model structure and training pipeline of the proposed \ourmethod{}. 
Instead of training a new model from scratch, we leverage existing large-scale diffusion models~\citep{rombach2022high} as a starting point.
Since these models have learned a strong image prior, it is more efficient to adapt the prior for a single image $p(\rvx)$ to model the joint distribution of a correlated image pair,  $p(\rvx, \rvy)$, than to learn this distribution from scratch.

Given a noisy image pair $(\rvx_{t_x}, \rvy_{t_y})$, we feed them as a batch into the denoising network.
They are simultaneously processed in two parallel branches, denoted as $\rvx$-branch and $\rvy$-branch.
By default, $\rvx$ is the image, and $\rvy$ is the condition.
When the conditional image differs from a natural image, we add a LoRA~\citep{hulora} to the $\rvy$-branch, which serves to adapt the image generator to produce, \eg depth or edges.
Additional joint cross-attention modules are injected parallel to the self-attention modules to join two branches. 
During training, we separately sample the timesteps $t_x, t_y$ for $\rvx, \rvy$ and optimize the model with the standard diffusion MSE loss from both branches.
Next we provide details on our joint cross-attention modules, LoRA adaptation, and training strategy.

\noindent \textbf{Joint cross attention.} The joint cross-attention module is the key component that enables ours model to learn a joint distribution $p(\rvx, \rvy)$ given the marginal distributions $p(\rvx), p(\rvy)$. It entangles $\rvx$-branch and $\rvy$-branch with cross branch attention.

The UNet~\citep{ronneberger2015u} is among the most common diffusion model implementation and consists of residual blocks and transformer blocks.
As shown in prior work~\citep{tumanyan2023plug}, the self-attention modules in the transformer blocks are crucial in determining the image structure and appearance.
Therefore, we inject the joint cross-attention modules in parallel to the self-attention modules. The module receives the features from both branches as input, with its outputs being added to the self-attention output of the two branches.
Specifically, given the input features of two branches $\mathbf{F}_{x}^\textrm{in}, \mathbf{F}_{y}^\textrm{in}$, the output features $\mathbf{F}_{x}^\textrm{out}, \mathbf{F}_{y}^\textrm{out}$ are computed as,
\begin{equation}
\begin{split}
    &\mathbf{F}_{x}^\textrm{joint}, \mathbf{F}_{y}^\textrm{joint} = \text{JointCrossAttn}(\mathbf{F}_{x}^\textrm{in}, \mathbf{F}_{y}^\textrm{in}), \\
    &\mathbf{F}_{x}^\textrm{out} = \text{SelfAttn}(\mathbf{F}_{x}^{in}) + \mathbf{F}_{x}^\textrm{joint},\quad \mathbf{F}_{y}^\textrm{out} = \text{SelfAttn}(\mathbf{F}_{y}^\textrm{in}) + \mathbf{F}_{y}^\textrm{joint}.
\end{split}
\label{eq:joint_cross_attn}
\end{equation}

In the joint cross-attention, two features $\mathbf{F}_{x}^\textrm{in}, \mathbf{F}_{y}^\textrm{in}$ attend to each other instead of attending to themselves as in the self-attention.
First, features are projected into queries $\mathbf{Q}_x, \mathbf{Q}_y$, keys $\mathbf{K}_x, \mathbf{K}_y$, and values $\mathbf{V}_x, \mathbf{V}_y$.
We use different matrices to project $\rvx$ features and $\rvy$ features.
For instance, $\mathbf{Q}_x = \mathbf{F}_{x}^{in} \mathbf{W}_{Q_x}, \mathbf{Q}_y = \mathbf{F}_{y}^{in} \mathbf{W}_{Q_y}$.
Then we perform cross-attention between $\rvx$ and $\rvy$ in bidirection,
\begin{equation}
    \mathbf{O}_{x} = \text{Softmax}(\frac{\mathbf{Q}_x\mathbf{K}_y^T}{\sqrt{d}})\cdot \mathbf{V}_y, \quad \mathbf{O}_{y} = \text{Softmax}(\frac{\mathbf{Q}_y\mathbf{K}_x^T}{\sqrt{d}})\cdot \mathbf{V}_x,
\end{equation}
where $d$ is the feature dimension and the feed-forward projection after the attention operation is omitted.
Essentially, $\rvx$ aggregates $\rvy$ values $\mathbf{V}_y$ according to the query-key similarity matrix $\mathbf{Q}_x\mathbf{K}_y^T$, and vice versa.
As a common practice~\citep{zhang2023adding, guo2023animatediff}, we add a zero-initialized linear projection ProjOut at the end to ensure the training starts without disrupting the pretrained feature distribution.
$\mathbf{F}_{x}^\textrm{joint}, \mathbf{F}_{y}^\textrm{joint} = \text{ProjOut}([\mathbf{O}_{x}, \mathbf{O}_{y}])$. $\mathbf{O}_{x}, \mathbf{O}_{y}$ are concatenated in channel dimension if image $\rvx$ and condition $\rvy$ are spatially-aligned to enhance feature fusion.
Otherwise, they are fed forward separately.

Compared to alternatives including feature residual~\citep{zhang2023adding}, input concatenation~\citep{stan2023ldm3d}, and backbone sharing~\citep{liu2023hyperhuman}, joint cross attention is compatible with image pairs without strict spatial correlation.
We initialize all joint cross-attention weights from the pretrained self-attention modules and train LoRA adapters for $\rvx,\rvy$ projection matrices.

\noindent \textbf{LoRA adaption for condition branch and joint cross attention.} Training all the parameters in our model for our conditional signal at least doubles the parameter number in pretrained image layers.
Therefore, we instead adopt the Low-Rank Adaptation technique~\citep{hulora} (LoRA) to fine-tune the pretrained weights by adding low-rank trainable weight matrices.
In addition to reducing trainable parameter numbers, using LoRA adapters allows us to apply our model to other checkpoints sharing the same structure as the training base model by plugging joint cross-attention and trained LoRA weights.

We freeze all pretrained layers in the $\rvx$-branch to retain the natural image prior $p(\rvx)$.
When the condition $\rvy$ is encoded as a pseudo-image falling out of natural image distribution, we add a LoRA adapter to the $\rvy$-branch to adapt for the condition image distribution $p(\rvy)$, denoted as $\rvy$-LoRA.
$\rvy$-LoRA applies to all projection matrices in the self-attention and cross-attention modules.

For joint cross-attention, we initialize all weights from the pretrained self-attention modules.
Then we add two sets of LoRA adapters to the pretrained projection matrices.
$\rvx\rvy$-LoRA includes $\mathbf{L}_{Q_x}, \mathbf{L}_{K_x}, \mathbf{L}_{V_x}$ and $\rvy\rvx$-LoRA includes $\mathbf{L}_{Q_y}, \mathbf{L}_{K_y}, \mathbf{L}_{V_y}$.
For instance, the adapted $\rvx, \rvy$ query projection matrices are $\mathbf{W}_{Q_x} = \mathbf{W}_{Q} + \mathbf{L}_{Q_x}, \mathbf{W}_{Q_y} = \mathbf{W}_{Q} + \mathbf{L}_{Q_y}$ where $\mathbf{W}_{Q}$ is the frozen query projection matrix from pretrained self-attention module.

\noindent \textbf{Training with disentangled noise levels.} One training objective adopted by previous methods~\citep{stan2023ldm3d, liu2023hyperhuman} is $\min_{\theta} E_{(\rvx_0, \rvy_0), \bm{\epsilon},t} \| \bm{\epsilon} - \bm{\epsilon}_{\theta} (\rvx_t,\rvy_t,t)\|^2$ where $\rvx,\rvy$ shares the same noise level.
The model learns how to denoise the noisy $(\rvx_t, \rvy_t)$ jointly and can generate new samples $(\rvx, \rvy) \sim p(\rvx, \rvy)$.
However, models trained in this way do not explicitly support conditional sampling.
Some work~\citep{zhang2023jointnet} solves the problem by augmenting the input with an condition mask and masked latents and finetuning the model with an inpainting target, yet this requires heavy training involving all model parameters.

Recently, Diffusion Forcing~\citep{chen2024diffusion} proposed a new training paradigm where the model is trained to denoise inputs with independent noise levels.
Inspired by the idea, we separately sample the diffusion timesteps for $\rvx$ and $\rvy$ when training, leading to the following training objective,
\begin{equation}
    \min_{\theta} E_{(\rvx_0, \rvy_0), \bm{\epsilon},t} \| \bm{\epsilon} - \bm{\epsilon}_{\theta} (\rvx_{t_x},\rvy_{t_y},t_x, t_y)\|^2,
\end{equation}
where $\bm{\epsilon} = (\bm{\epsilon}_x, \bm{\epsilon}_y)$ and
$\rvx_{t_x} = \text{AddNoise}(\rvx_0, t_x), \rvx_{t_y} = \text{AddNoise}(\rvy_0, t_y)$.
Models trained with the timestep-disentangled objective can directly perform conditional generation by denoising $\rvx_{t_x}$ while keeping $t_y=0$.
The noise added to the input can be regarded as an implicit noise mask. Unlike explicit input masks, noise masking can interpolate between no mask ($t=0$) and full mask $t=T$, enabling image generation with coarse conditions.

\subsection{Inference}
\label{sec:infer}
Our timestep-disentangled training allows \ourmethod{} models to process paired inputs with different noise levels in each denoising step.
Suppose we have a denosing sequence $\{(\rvx_{i}, \rvy_{i})\}_{i=S}^{0}$. $\rvx_{i}, \rvy_{i}$ are sampled under independent noise schedules $(t_x^{S},...,t_x^{0})$ and $(t_y^{S},...,t_y^{0})$ where $t^{i} \leq t^{i+1}$.

\noindent \textbf{Sampling with independent noise schedules.}
The independent $\rvx,\rvy$ noise schedule enables diverse sampling behaviors.
First, we can jointly generate $\rvx,\rvy$ by denoising them together with identical noise schedules, $t^i_y = t^i_x$ $\forall i$.
For conditional generation, we can sample $\rvx$ from noise with a clean condition input $\rvy$, \ie $\rvy_{i}=\rvy, t^i_y=0$ $\forall i$.
We can similarly sample $\rvy$ conditioned on $\rvx$ by giving the $\rvx$-branch clean input and $t^i_x=0$ $\forall i$.
Furthermore, our models allow sampling $\rvx$ with a coarse control signal conditioning on a noisy condition image $\rvy_{S}=\text{AddNoise}(\rvy, t^S_y)$.
We can control the condition fidelity by adjusting the noise level $t^0_y$ from $0$ (no noise) to $T$ (pure noise).
Since the control signal is corrupted by noise, the condition image itself does not need to be precise.
Therefore, we can use artificially created or edited condition images to loosely control image generation.

\noindent \textbf{Sampling with guidance.} Since our model has an output for both branches, we can apply guidance to each of them for image inpainting or partial condition. Latent replacement is a typical approach for inpainting where the noisy latents $\rvz_t$ are partially replaced by exact samples from the forward process (Eq.~\ref{eq:diff_forward}) in each step, $\rvz_t=(1-\vm) \cdot \text{AddNoise}(\rvz, t) + \vm \cdot \rvz_t$ where $\rvz,\vm$ are the given condition sample and mask. The method is an approximation to exact conditional sampling.
Following~\citet{ho2022video}, we can add a guidance term to correct the sampling process and improve the condition adherence,
\begin{equation}
    \rvz_t^{g} = \rvz_t - w_r \frac{\bar{\alpha}_t}{2} \nabla_{\rvz_t^{\tilde{m}}} || \rvz^{m} - \hat{\rvz}^{m}_{0}(\rvz_t;\theta)||^2,
    \label{eq:guidance}
\end{equation}
where $\rvz_t^{g}$ is guided noisy latents, $\hat{\rvz}_{0}(\rvz_t;\theta) = [\rvz_t - \sqrt{1-\bar{\alpha}_t} \bm{\epsilon}_{\theta}(\rvz_t,t)] / \sqrt{\bar{\alpha}_t}$ is the predicted original sample and $w_r$ is a weighting factor. $\rvz^{m}$ indicates part of $\rvz$ where mask $\vm=0$ and $\tilde{\vm}$ is the inverted mask.
The guidance term leads the noisy latents toward reconstructing the masked condition area $\rvz^{m}$. For \ourmethod{}, above variables include both inputs, $\rvz_t=(\rvx_{t_x},\rvy_{t_y}), \rvz=(\rvx,\rvy),\vm=(\vm_x,\vm_y)$.

\noindent \textbf{Sampling with multiple conditional signals.} To sample with multiple conditional images, we combine multiple \ourmethod{} models and extend the joint cross attention to include all image-condition pairs. In specific, the image feature is paired with each condition feature in the joint cross-attention modules and processed by weights from corresponding models. Then the image branch aggregates all output features with weight factors to balance the strength of each condition.
\section{Results}
\label{sec:results}

\begin{figure}[t]
\begin{center}
\includegraphics[width=0.9\linewidth]{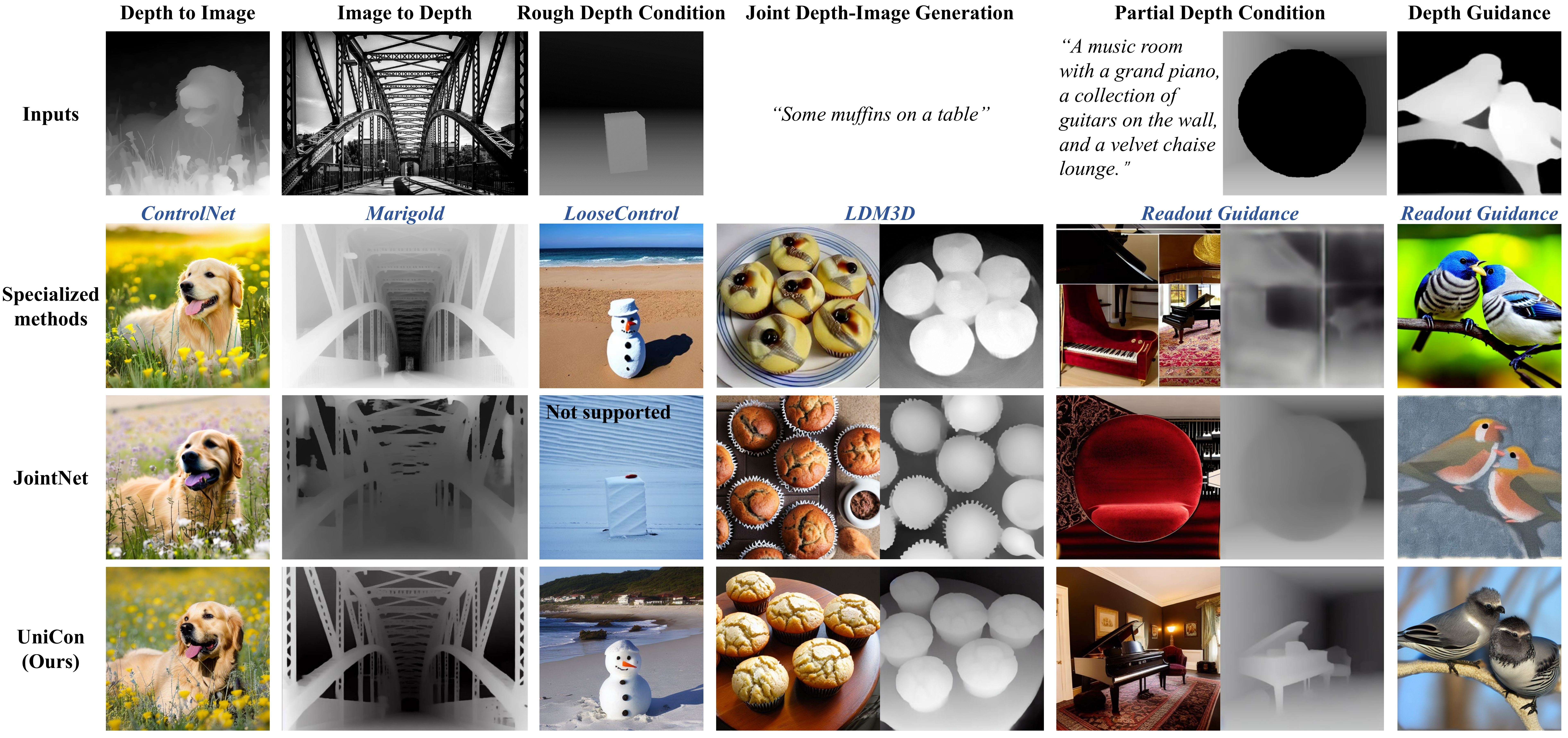}
\end{center}
\vspace{-0.5cm}
\caption{\textbf{Qualitative comparison of diverse Image-Depth generation tasks.} We compare our single \ourmethod{}-Depth model with other specialized methods and a previous unified method JointNet~\citep{zhang2023jointnet} on diverse generation tasks.
}
\label{fig:main_comp}
\vspace{-0.3cm}
\end{figure}

\begin{figure}[t]
\begin{center}
\includegraphics[width=0.8\linewidth]{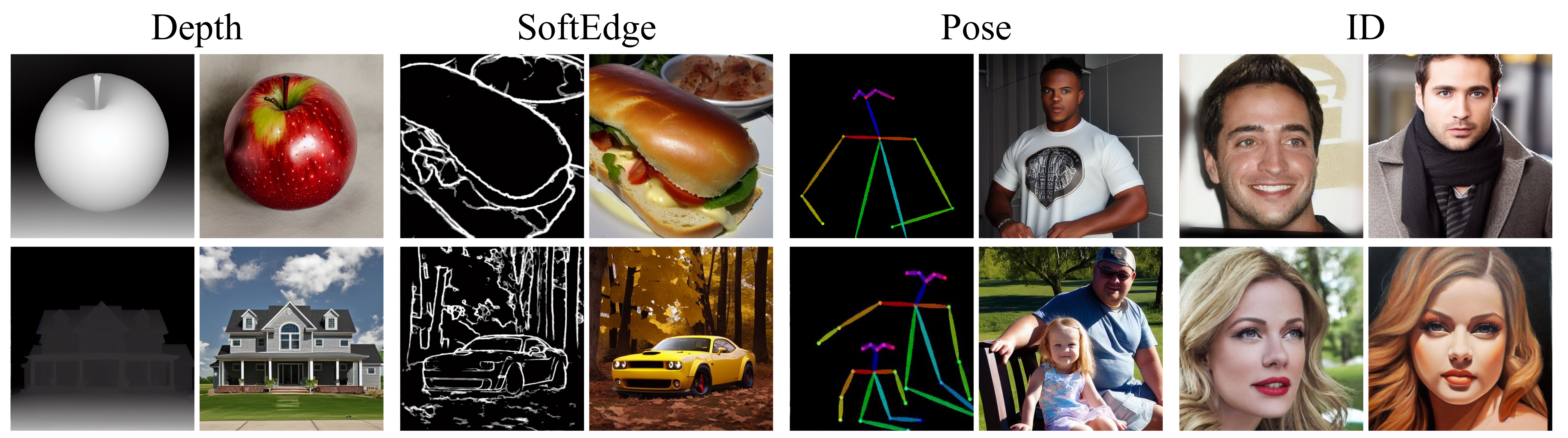}
\end{center}
\vspace{-0.3cm}
\caption{\textbf{Conditional generation samples}. We show our sample conditional generation results. For each model, the left column is the input condition and the right column is the output image.
}
\vspace{-0.3cm}
\label{fig:sample_res}
\end{figure}

We base our experiments on Stable Diffusion~\citep{rombach2022high} (SD), a large-scale text-to-image diffusion model. We train 4 \ourmethod{} models, Depth, SoftEdge, Human-Pose (Pose), Human-Identity (ID) on SDv1-5. The first three pair an image with a spatially aligned condition image. Following~\citet{luo2024readout}, we train Depth, SoftEdge models on 16k images from PascalVOC~\citep{pascal-voc-2012} and Pose model on a subset with 9k human images. Depth, soft edge, and pose images are estimated by Depth-Anything-v2~\citep{depth_anything_v2}, HED~\citep{xie2015holistically} and OpenPose~\citep{cao2017openpose}. We train the ID model on 30k human face images from CelebA~\citep{liu2015faceattributes} and use images with the same identity as training image pairs.
In addition, we train an auxiliary Appearance model to cooperate with other models, as shown in Fig~\ref{fig:multi-sig}. It is trained on random frame pairs from 6k videos in Panda70M~\citep{chen2024panda}, aiming at generating images with similar visual appearance.

\subsection{Main results}

\noindent \textbf{Qualitative results.} In Fig.~\ref{fig:main_comp}, we show sample results generated by our Depth model on different tasks and compare them with a specific method for each task. Note our results are generated by the same Depth model with different sampling strategy. First, our model can accept a clean depth or image to perform Depth-to-Image or Image-to-Depth generation. ControlNet~\citep{zhang2023adding} and Marigold~\citep{ke2023repurposing} respectively work for the two tasks with different structures. Compared to ControlNet, \ourmethod{} supports generation with a rough or partial condition. LooseControl~\citep{bhat2024loosecontrol} finetunes a ControlNet for a generalized depth condition. Our model works for such created rough depth images without foreknowledge by conditioning noisy depth images. Similar to Readout-Guidance~\citep{luo2024readout}, we can apply the guidance on the depth output for conditional generation. In addition to a complete depth image, it is also possible to guide with part of a depth image, such as using the border of the depth map to specify the overall scene structure. Finally, our model can jointly generate an image with its depth, which is the goal of LDM3D~\citep{stan2023ldm3d}.

Fig~\ref{fig:sample_res} shows sample results from all \ourmethod{} models. Apart from common spatial aligned control signals, our ID model works for loosely correlated human images. With the ID model, we can generate images of the same person in the condition image and utilize the input prompt to specify appearance and style.

\begin{figure}[t]
\begin{center}
\includegraphics[width=\linewidth]{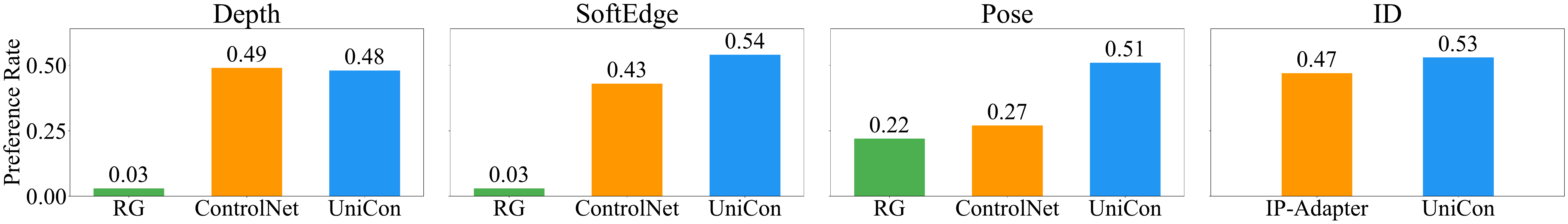}
\end{center}
\vspace{-0.3cm}
\caption{\textbf{User study of conditional generation performance.} We compare our \ourmethod{} against Readout-Guidance~\citep{luo2024readout}, ControlNet~\citep{zhang2023adding} for spatially-aligned conditions (depth, softedge, pose) and against IP-Adapter-Face~\citep{ye2023ip} for ID condition.
}
\vspace{-0.3cm}
\label{fig:user-study}
\end{figure}

\begin{table}[t]\scriptsize
\caption{\textbf{Quantitative comparison of conditional generation for spatially-aligned conditions.} We compare  Readout-Guidance~\citep{luo2024readout}, ControlNet~\citep{zhang2023adding}, and \ourmethod{} (Ours) on spatially-aligned conditions depth, softedge, and pose.
For a fair comparison, all methods are trained on PascalVOC with the same annotator.
}
\label{tab:main_res}
\begin{center}
\vspace{-0.3cm}
\begin{tabular}{lcccccc}
\toprule
\multirow{2}{*}{Method} & \multicolumn{2}{c}{Depth} & \multicolumn{2}{c}{SoftEdge} & \multicolumn{2}{c}{Pose}
\\
  & \multicolumn{1}{c}{\bf FID-6K $\downarrow$} & \multicolumn{1}{c}{\bf AbsRel (\%) $\downarrow$} & \multicolumn{1}{c}{\bf FID-6K $\downarrow$} & \multicolumn{1}{c}{\bf EMSE (1e-2) $\downarrow$} & \multicolumn{1}{c}{\bf FID-6K $\downarrow$} & \multicolumn{1}{c}{\bf PCK @ 0.1 $\uparrow$}
 \\
 \midrule
Readout-Guidance        &  18.72 &  23.19 & 18.43 & 4.84 & 21.07 & 24.96\\
ControlNet             &  13.68 &  9.85 & 13.46 & 2.30 & 18.61 & 57.54\\
\ourmethod{}            & \textbf{13.21} &  \textbf{9.26} & \textbf{12.80} & \textbf{2.28} & \textbf{17.51} & \textbf{61.97}\\
\bottomrule
\end{tabular}
\vspace{-0.6cm}
\end{center}
\end{table}

\begin{wrapfigure}[15]{R}{.42\textwidth}
\vspace*{-4mm}
\scriptsize\centering
\captionof{table}{\textbf{Quantitative depth estimation comparison.} We compare MiDaS~\citep{ranftl2020towards}, DPT~\citep{ranftl2021vision}, Marigold~\citep{ke2023repurposing}, and our Depth-Metric model on zero-shot depth estimation benchmarks. We show results without test-time ensembling.}
\label{tab:estimation_res}
\begin{tabular}{@{}lcccc@{}}
\toprule
\multirow{2}{*}{Method} & \multicolumn{2}{c}{NYUv2} & \multicolumn{2}{c}{ScanNet}
\\
  & \multicolumn{1}{c}{\bf AbsRel $\downarrow$} & \multicolumn{1}{c}{$\mathbf{\delta1}$ $\uparrow$} & \multicolumn{1}{c}{\bf AbsRel $\downarrow$} & \multicolumn{1}{c}{\bf $\mathbf{\delta1}$ $\uparrow$}
 \\
 \midrule
MiDaS        &   11.1 & 88.5 & 12.1 & 84.6\\
DPT         &  9.8 & 90.3 &  8.2 & 93.4\\
Marigold             & 6.0 & 95.9 & 6.9 & 94.5 \\
\midrule
JointNet & 13.7 & 81.9 & 14.7 & 79.5 \\
\ourmethod{}            & 7.9 & 93.9 & 9.2 & 91.9\\
\bottomrule
\end{tabular}
\end{wrapfigure}
\noindent \textbf{Quantitative comparison.} We compare \ourmethod{} with other methods on conditional generation and depth estimation. For the conditional generation, we generate 6K $512\times512$ images conditioned on depth, soft edge, or pose of random images from OpenImages~\citep{openimages}. We use Frechet Inception Distance (FID)~\citep{heusel2017gans} to measure the distribution distance between generated images and real images corresponding to the same input conditions. We also evaluate the condition fidelity by Absolute Mean Relative Error (AbsRel)~\citep{ke2023repurposing} for depth; Edge Mean Squared Error (EMSE) for soft edge; and Percentage of Correct Keypoints (PCK)~\citep{yang2012articulated} for pose. All metrics are computed between the modalities estimated from real images and generated images. For depth estimation, we evaluate on NYUv2~\citep{silberman2012indoor} and ScanNet~\citep{dai2017scannet} with AbsRel and $\delta1$~\citep{ranftl2021vision} as metric, following the protocol of affine invariant depth evaluation~\citep{ranftl2020towards}.

We conduct a user study to compare \ourmethod{} models against their corresponding methods to evaluate the conditional generation performance. As shown in Fig.~\ref{fig:user-study}, \ourmethod{} demonstrates comparable performance to specialized methods in human preference (Fig.~\ref{fig:user-study}).

In addition,
we compare \ourmethod{} for spatially-aligned conditions (depth, softedge, pose) against ControlNet~\citep{zhang2023adding}, Readout-Guidance~\citep{luo2024readout} in terms of FID and condition fidelity metrics.
As shown in Tab.~\ref{tab:main_res}, our \ourmethod{} achieves similar or better performance than Readout-Guidance~\citep{luo2024readout} and ControlNet~\citep{zhang2023adding} on FID and condition fidelity over all modalities.

We finetune our Depth model for 5K steps with Depth-Anything-V2-Metric~\citep{depth_anything_v2} as the annotator for metric depth evaluation. As shown in Tab.~\ref{tab:estimation_res}, our Depth-Metric model performs similarly or better than MiDaS~\citep{ranftl2020towards} and DPT~\citep{ranftl2021vision}. There is a margin between our model and Marigold~\citep{ke2023repurposing}, which fine-tunes the whole diffusion model and solely focuses on depth estimation. In comparison, our model trains fewer parameters and targets on unified conditional generation.

\begin{figure}[t]
\begin{center}
\includegraphics[width=\linewidth]{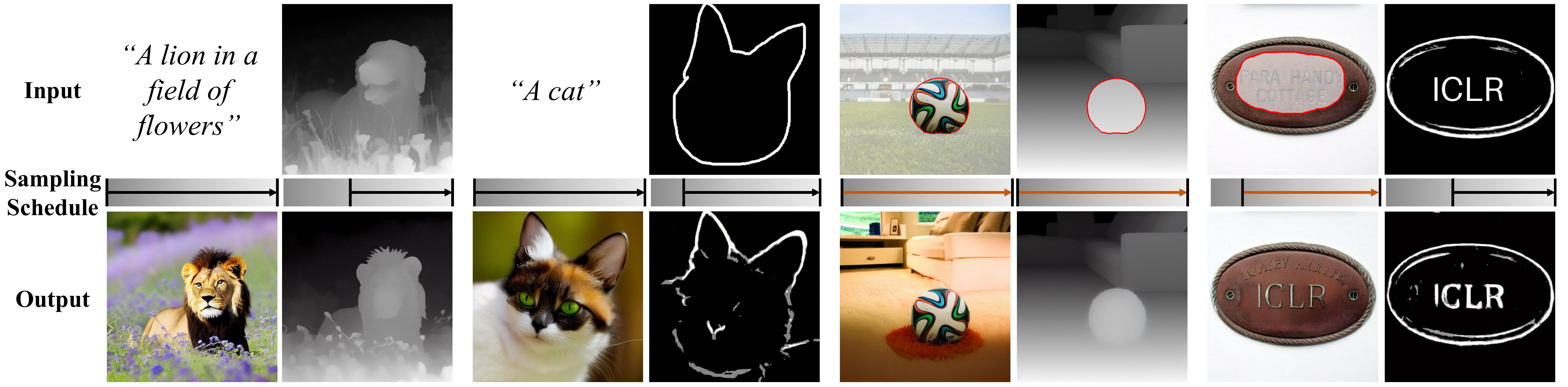}
\end{center}
\vspace{-0.3cm}
\caption{\textbf{Flexible conditional generation via different sampling schedules.} We annotate each image with its sampling schedule. Schedule bars represent the noise level from $T$ to $0$. Arrows indicate the noise sampling schedule. We apply the guidance in generation (orange arrow) for partial condition samples and use red borderlines to split areas to keep and areas to generate.
}
\label{fig:flexible_cond}
\vspace{-0.3cm}
\end{figure}

\begin{figure}[t]

\begin{center}

\includegraphics[width=\linewidth]{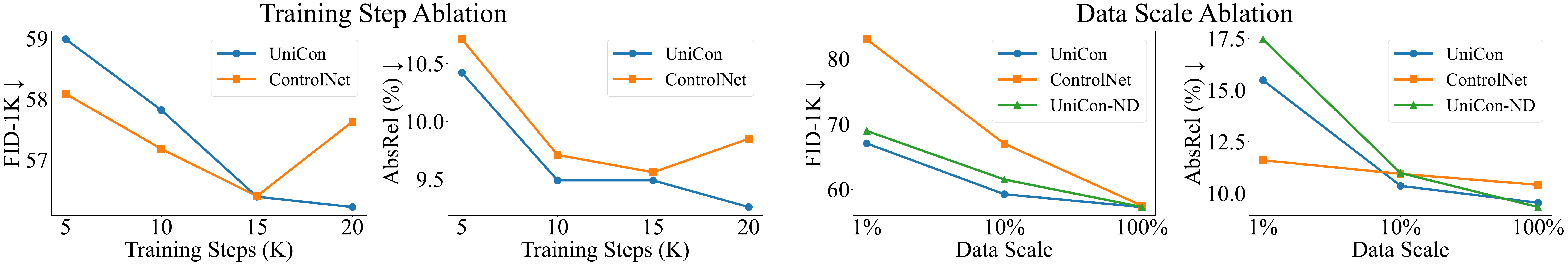}
\end{center}

\vspace{-0.3cm}

\caption{\textbf{Ablation of training steps and data scale.} We compare the depth conditional generation performance of \ourmethod{} and ControlNet on different training settings. We use the same training steps for data scale ablation and vice versa. \ourmethod{}-ND indicates trained without depth loss.
}
\label{fig:ablations}
\vspace{-0.6cm}
\end{figure}

\noindent \textbf{Flexible conditional generation.} We show diverse conditional generation samples in Fig.~\ref{fig:flexible_cond}. Starting from a noisy condition image, our models can interpret the exact condition image to other meanings (dog to lion in Column 1) or take rough condition as control (cat sketch in Column 2). Using guidance for partial conditioning enables us to condition on both input signals (Column 3) or repaint an image with a coarse condition signal (Column 4). We generate the "ICLR" edge image by replacing raw edges with text.

\begin{figure}[t]
\begin{minipage}{0.45\textwidth}
\begin{center}
\includegraphics[width=0.9\linewidth]{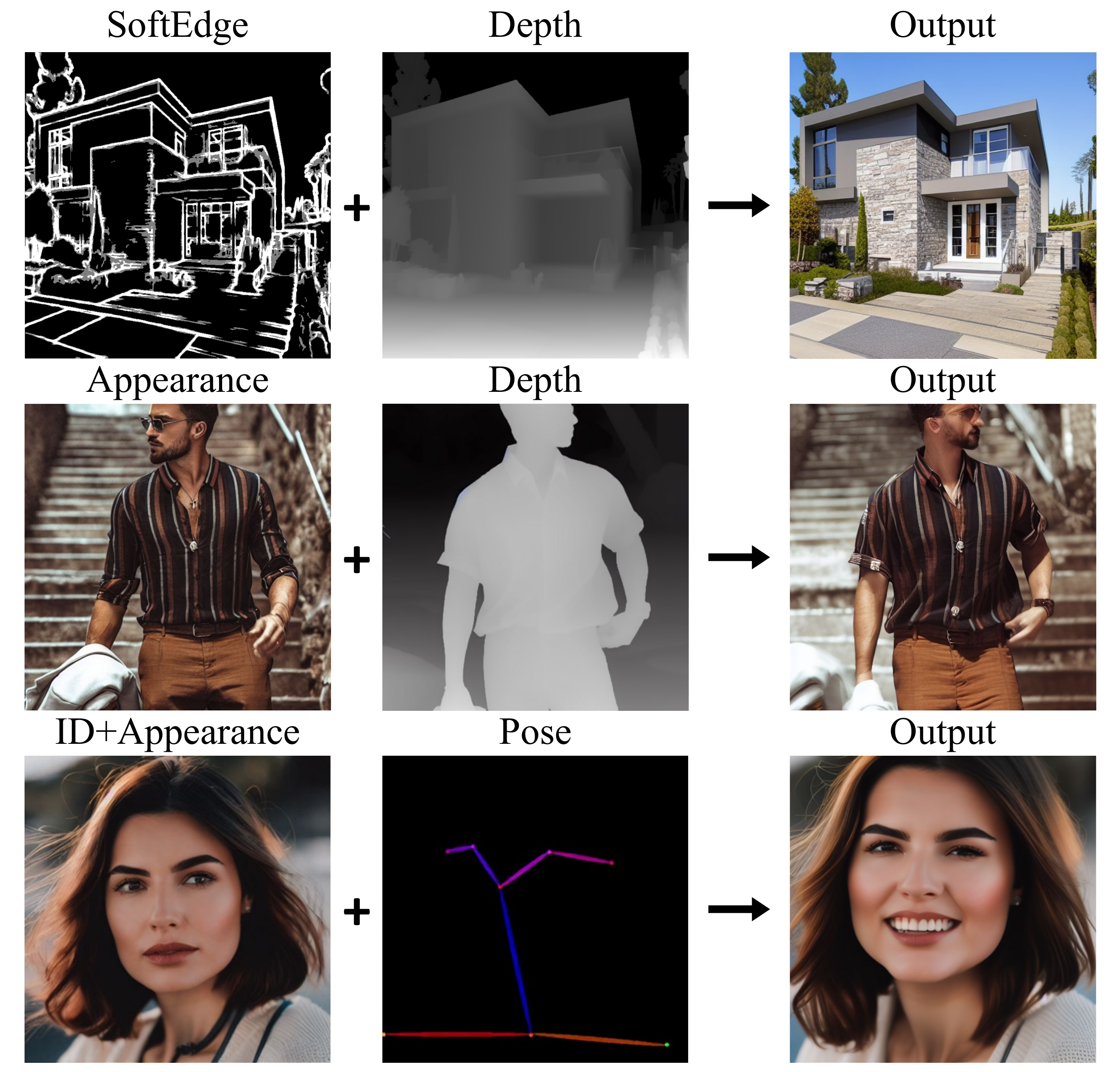}
\end{center}
\caption{\textbf{Multi-signal conditional generation samples.} We can combine multiple models for different condition signals.}
\label{fig:multi-sig}
\end{minipage}
\hfill
\begin{minipage}{0.51\textwidth}
\begin{center}
\includegraphics[width=\linewidth]{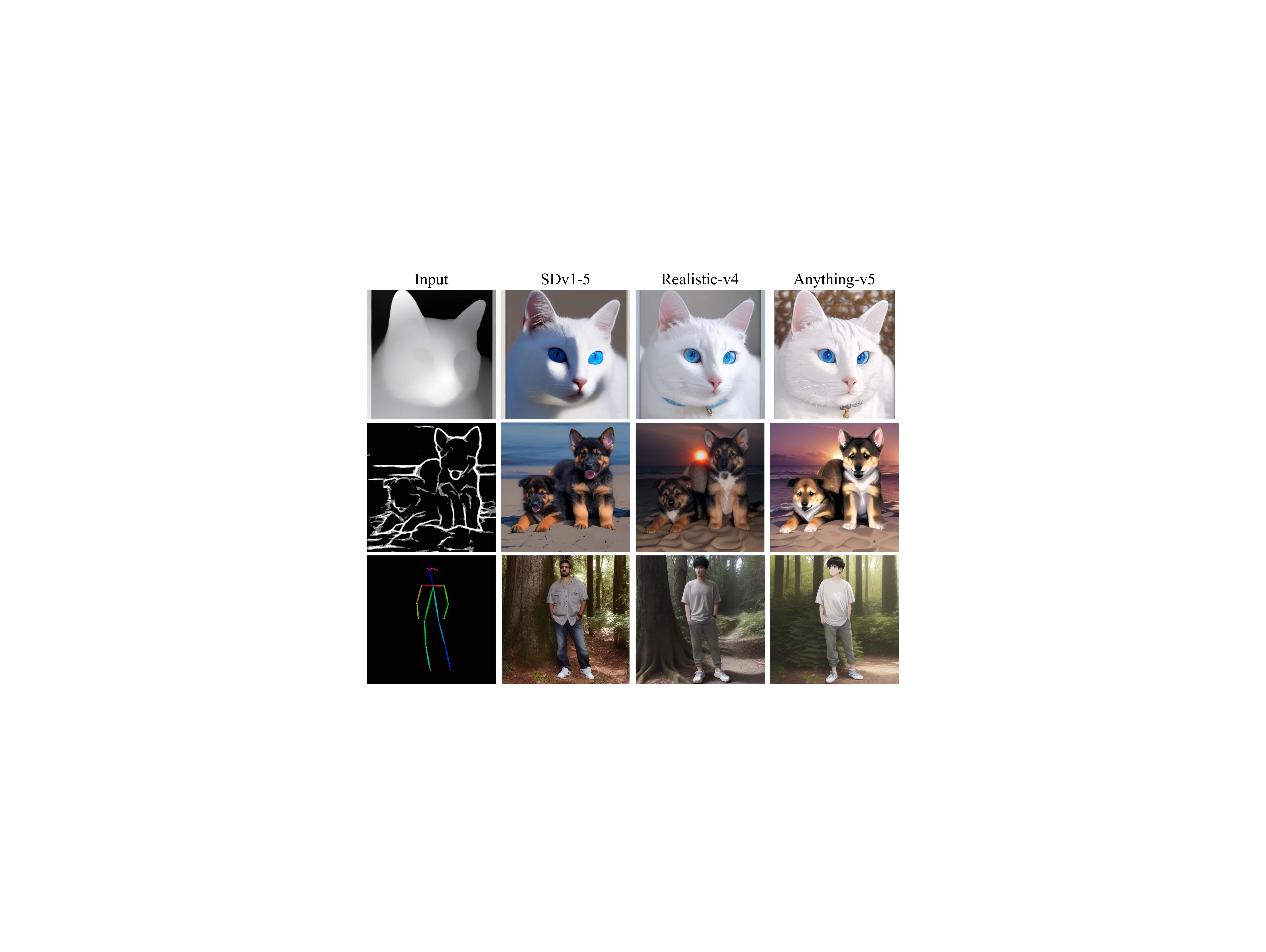}
\end{center}
\caption{\textbf{Apply to other checkpoints.} Our models can apply to other checkpoints fine-tuned from the base model (SDv1.5).}
\label{fig:switch-base}
  \end{minipage}
\vspace{-0.9cm}
\end{figure}

In addition, we can combine multiple \ourmethod{} models to enhance the control ability (Fig.~\ref{fig:multi-sig}). One interesting application is to combine loose conditions (ID, Appearance) with dense conditions (Depth, Pose). In the bottom row, we use the same image for both ID and Appearance conditions to enhance both ID alignment and overall image appearance consistency.
Similar to ControlNet, our models can apply to other customized checkpoints fine-tuned from our base model (Fig.~\ref{fig:switch-base}).

\noindent \textbf{Comparison with JointNet.} We compare our method to the most relevant baseline, JointNet~\citep{zhang2023jointnet}, which also supports multiple conditional generation tasks for image and depth. As demonstrated in Fig.~\ref{fig:main_comp}, our method shows better results in image-to-depth, partial condition, and depth guidance, with extra abilities of rough depth condition. Quantitatively, \ourmethod{} achieves superior depth estimation results compared to JointNet (Tab.~\ref{tab:estimation_res}). 



\subsection{Ablation Study}

\noindent \textbf{Training steps and data scale.} We ablate on training steps and data scale for depth conditional generation (Fig.~\ref{fig:ablations}). As the training step grows, our model improves steadily, showing a higher performance upper bound than ControlNet which starts to overfit on the dataset. We observe a sudden drop in condition fidelity when downscaling our training dataset. We attribute it to our joint cross-attention requiring a certain data scale to capture the image-condition correlation. On the other hand, \ourmethod{} achieves better condition fidelity with enough data.

\begin{wrapfigure}[12]{R}{.37\textwidth}\scriptsize
\vspace*{-4mm}
\captionof{table}{\textbf{Ablation of training setting and model alternatives.} We evaluate the conditional generation performance of our Depth model under different settings.}
\label{tab:abl}
\vspace*{-3mm}
\begin{center}
\begin{tabular}{@{}lcc@{}}
\toprule
& \bf FID-6K $\downarrow$ & \bf AbsRel(\%) $\downarrow$
 \\
 \midrule
\ourmethod{}-Depth & 13.21 & 9.26 \\
- Depth loss & 13.18 & 9.23 \\
- Depth loss, noise & 13.66 & 8.57 \\
- Encoder & 13.64 & 10.16 \\
+ Data (200K) & 13.10 & 8.66 \\
\bottomrule
\end{tabular}
\end{center}
\end{wrapfigure}
\noindent \textbf{Training setting and model alternatives.}
In Tab.~\ref{tab:abl}, we test our Depth model with training setting and structure alternatives.
First, we drop out the depth loss in training, leading to a model that has similar depth-to-image generation performance but cannot denoise depth image. We also investigate the depth loss influence under different data scales (Fig.~\ref{fig:ablations} Right), showing that joint modeling depth has a positive effect on conditional generation. Further dropping the noise added to the depth image in training results in a ControlNet-like depth-control model. It improves the condition fidelity but harms generation quality. We also remove joint cross-attention modules in the UNet encoder to test the robustness against structure changes. Despite the slight performance drop, our method works with half of the attention modules. Finally, our model consistently improves when we scale up the training data to 200K images from OpenImages~\citep{openimages}.

\begin{figure}[t]

\end{figure}
\section{Discussion}
\label{sec:discussion}

We propose a simple framework for unifying diffusion-based conditional generation. We consider all conditional generation tasks involving a specific image-condition correlation as sampling in a global distribution and train a diffusion model to learn it. Our flexible model architecture adapts a pretrained diffusion model to handle multi-input processing, alongside effective training and sampling strategies designed to support diverse generation tasks. The inherent flexibility of our approach opens up the possibility for a wide range of applications, potentially encouraging further exploration into novel image-condition mappings. Additionally, our work demonstrates that large-scale diffusion models can be successfully adapted to accommodate non-aligned noise levels in input signals, suggesting a path toward enhancing existing multi-signal diffusion models. As for limitations, some models dealing with loosely correlated image pairs, \eg our ID model, exhibit instability. We attribute this issue to the need for more training data and refined techniques to achieve satisfactory performance in such cases.

\noindent \textbf{Acknowledgements.} This work was supported in part by the National Natural Science Foundation of China (62322113, 62376156).

\bibliography{iclr2025_conference}
\bibliographystyle{iclr2025_conference}

\appendix
\section{Implementation Details}
Our model process input $\rvx, \rvy$ in two parallel branches, as shown in Fig.~\ref{fig:pipeline}. In practice, we do not prepare two network branches to process two inputs. Instead, all inputs are concatenated in batch dimension and fed into the denoising UNet, as a batch of image inputs. They are simultaneously processed, with the condition LoRA selectively applying to the $\rvy$ inputs. For joint cross-attention, we split $\rvx$ and $\rvy$ features and perform the cross-attention operation.

For the models we present in the paper, we use LoRA rank 64 for all adapters, including the condition LoRA and the joint cross-attention LoRA. We only add the condition LoRA when the condition image falls out of natural image distribution, \ie our Depth, SoftEdge, and Pose models. We additionally incorporate a trigger word to the text prompts for these conditions, such as "depth\_map".

\section{Training}
We train 5 \ourmethod{} models, Depth, SoftEdge, Human-Pose (Pose), Human-Identity (ID), and Appearance on Stable Diffusion v1.5~\citep{rombach2022high}.
Stable Diffusion uses a variational autoencoder (VAE) to define a latent space for image generation. We adopt the VAE of the base model to encode and decode all type of images to and from the latent space, including annotated images like depth, edge and pose.
For all models, we use AdamW~\citep{loshchilov2017decoupled} optimizer with learning rate 1e-4. The training images are resized to 512 resolution with random flipping and random cropping as data augmentation.
The text prompts are generated by BLIP~\citep{li2023blip,li2022blip} for datasets without captions.
We drop out the text prompt input with a rate of 0.1 to maintain the classifier-free guidance~\citep{ho2022classifier} ability.

\noindent \textbf{Depth, SoftEdge, Pose.}
For spatially aligned conditions, we follow Readout-Guidance~\citep{luo2024readout} to train on PascalVOC~\citep{pascal-voc-2012}. Depth and SoftEdge model is trained on 16K images and Pose model is trained on 9K images of humans. To obtain the condition input, we first annotate training images with existing estimation methods. Depth, soft edge, and pose images are estimated by Depth-Anything-v2~\citep{depth_anything_v2}, HED~\citep{xie2015holistically} and OpenPose~\citep{cao2017openpose}.
We encode all estimated modalities as images, following the annotators in ControlNet~\citep{zhang2023adding}.
For SoftEdge, we follow ControlNet~\citep{zhang2023adding} to quantize the edge maps into several levels to remove possible hidden patterns. We train Depth, SoftEdge models for 20K steps with batch size 32 and Pose model for 10K steps. Training 20K steps costs about 13 hours on two NVIDIA A800 80G GPUs. To adapt our Depth model for metric depth estimation, we further fine-tune the model for 5K steps on images annotated with Depth-Anything-v2-Metric. The produced Depth-Metric model is used for the depth estimation evaluation in Tab.~\ref{tab:estimation_res}.

\noindent \textbf{ID, Appearance.}
We test our method on two cases of loosely correlated image pairs, the identity-preserving model and the appearance-preserving model. The Identity model is trained on human images with the same identity.
We collect 30K human images from CelebA~\citep{liu2015faceattributes}, including about 5K identities with more than 1 image. We randomly pair images with the same ID to generate 200K training image pairs (50 for each identity). 
We train the Appearance model on Panda-70M~\citep{chen2024panda}. Due to issues in downloading YouTube videos in Panda-70M, we only use a minimal subset with 6K videos for training.
When training, we load video clips with a length of 16 and a resampled frame rate of 7 and randomly select two frames from loaded clips as input image pairs.
The ID and Appearance models are trained for 20k steps with batch size 64 distributed on 4 NVIDIA A800 80G GPUs.

\section{Inference}
\begin{table}[htbp]
\vspace{-0.1cm}
\caption{Example sampling schedules for different conditional generation targets given a \ourmethod{} model of $\rvx,\rvy$. }
\label{tab:sample_schedule}
\begin{center}
\vspace{-0.2cm}
\begin{tabular}{lc}
\toprule
Target & Example Sampling Schedule (50 steps)
 \\
 \midrule
$p(\rvx,\rvy)$ & $(\rvx_{50}, \rvy_{50}), (\rvx_{49}, \rvy_{49}), ..., (\rvx_{1}, \rvy_{1}), (\rvx_{0}, \rvy_{0})$ \\
$p(\rvy|\rvx_0)$ & $(\rvx_0, \rvy_{50}), (\rvx_0, \rvy_{49}), ..., (\rvx_0, \rvy_{1}), (\rvx_0, \rvy_{0})$ \\
$p(\rvx|\rvy_0)$ & $(\rvx_{50}, \rvy_{0}), (\rvx_{49}, \rvy_{0}), ..., (\rvx_{1}, \rvy_{0}), (\rvx_{0}, \rvy_{0})$ \\
$p(\rvx|\rvy_{25})$ & $(\rvx_{50}, \rvy_{25}), (\rvx_{49}, \rvy_{25}),(\rvx_{48}, \rvy_{24}), ..., (\rvx_{1}, \rvy_{1}), (\rvx_{0}, \rvy_{0})$ \\
$p(\rvx|\rvy^{m}_{0})$ & $(\rvx_{50}, \rvy_{50}), (\rvx^{g(y^{m}_{0})}_{49}, \rvy^{g(y^{m}_{0})}_{49}), ..., (\rvx^{g(y^{m}_{0})}_{1}, \rvy^{g(y^{m}_{0})}_{1}), (\rvx^{g(y^{m}_{0})}_{0}, \rvy^{g(y^{m}_{0})}_{0})$\\
\bottomrule
\end{tabular}
\vspace{-0.3cm}
\end{center}
\end{table}

\noindent \textbf{Sampling schedules}. In Tab~\ref{tab:sample_schedule}, we list example sampling schedules to show how our sampling strategies (discussed in Sec.~\ref{sec:infer}) support the conditional sampling described in Eq.~\ref{eq:joint_cond_distribution}. For the partial conditioning case $p(\rvx|\rvy^{m}_0)$, $g(\rvy^{m}_0)$ means replacing latents and applying guidance according to give masked condition $\rvy^{m}_0$ (Sec.~\ref{sec:infer}, Sampling with guidance). Note that the listed sampling schedule is one possible schedule to achieve the target. We can alter or combine them to perform customized conditional generation. For guidance, we adopt an optimizer (\eg AdamW) to compute the gradient and determine the weighting factor $w_r$ in Eq.~\ref{eq:guidance} instead of manually setting a fixed weight factor, as suggested by Readout-Guidance~\citep{luo2024readout}.

\noindent \textbf{Combining multiple models}. As discussed in Sec.~\ref{sec:infer}, we can combine multiple \ourmethod{} models to achieve multi-signal control. We expand the details here. Suppose we have input image $\rvx$ and two input conditions $\rvy, \rvz$. We denote the model parameters for $\rvy$ and $\rvz$ as $\theta_{y}, \theta_z$. Then the joint feature outputs $\mathbf{F}^{joint}_{x}, \mathbf{F}^{joint}_{y}, \mathbf{F}^{joint}_{z}$ in Eq.~\ref{eq:joint_cross_attn} are computed as:

\begin{equation}
\begin{split}
    &\mathbf{F}_{xy}, \mathbf{F}_{yx} = \text{JointCrossAttn}(\mathbf{F}_{x}^\textrm{in}, \mathbf{F}_{y}^\textrm{in};\theta_{y}),
\mathbf{F}_{xz}, \mathbf{F}_{zx} = \text{JointCrossAttn}(\mathbf{F}_{x}^\textrm{in}, \mathbf{F}_{z}^\textrm{in};\theta_{z}),\\
&\mathbf{F}^{joint}_{x} = w_{xy}\mathbf{F}_{xy} + w_{xz}\mathbf{F}_{xz}, \mathbf{F}^{joint}_{y} = w_{yx}\mathbf{F}_{yx}, \mathbf{F}^{joint}_{z} = w_{zx}\mathbf{F}_{zx},
\end{split}
\label{eq:multi_sig_cross_attn}
\end{equation}
where all $w$ are weighting factors to balance the strength of each condition. To explain, we perform joint cross-attention between any image-condition pairs $(\rvx,\rvy)$ and $(\rvx,\rvz)$ with corresponding model weights. Then the image branch will aggregate all output features as the final output.

\noindent \textbf{Condition Guidance.} Our ID and Appearance models that target a loose correlation sometimes perform badly in the conditional generation, generating low-quality images or images not aligned with the condition. We attribute the problem to the fact that loose condition has a weaker influence on image generation as it allows more freedom and diversity in generated images than dense conditions. Similar problems have also been observed in text-to-image generation, with an effective solution called classifier-free guidance~\citep{ho2021classifier}. Therefore, we optionally utilize a similar guidance scheme to emphasize a certain condition signal for better condition alignment.

In specific, we alter the model output as $\bm{\epsilon}_g = \bm{\epsilon}_{sep} + k (\bm{\epsilon}_{joint} - \bm{\epsilon}_{sep})$ where $\bm{\epsilon}_{joint}, \bm{\epsilon}_{sep}$ are the model output with or without joint cross-attention and $k$ is the guidance scale.
Intuitively, the output towards the direction defined by $\bm{\epsilon}_{joint} - \bm{\epsilon}_{sep}$, which means the condition signal from joint cross-attention is enhanced.
Furthermore, we can enable more fine-grained guidance over a specific signal when there are multiple condition signals by replacing $\bm{\epsilon}_{joint}, \bm{\epsilon}_{sep}$ with $\bm{\epsilon}(\vw_{con}), \bm{\epsilon}(\vw_{unc})$. Here $\vw_{con},\vw_{unc}$ indicate the two sets of weighting factors used in Eq.~\ref{eq:multi_sig_cross_attn}. Therefore $\bm{\epsilon}(\vw_{con}), \bm{\epsilon}(\vw_{unc})$ are the model output with different condition weighing factors. Take Eq~\ref{eq:multi_sig_cross_attn} as an example, if we want to emphasize $\rvy$ but not $\rvz$, we can set $w_{xy},w_{yx}$ to 1 in $\vw_{con}$ and to 0 in $\vw_{unc}$ while keeping other weights the same.

\section{Evaluation}
\begin{table}[t]\scriptsize
\caption{\textbf{Relative depth estimation performance on DA-2K evaluation benchmark.} We compare \ourmethod{}-Depth model against Marigold~\citep{ke2023repurposing}, Geowizard*~\citep{fu2025geowizard}, Depth Anything V1~\citep{depth_anything_v1} (DA V1), ZoeDepth~\citep{bhat2023zoedepth}, Depth Anything V2~\citep{depth_anything_v2} (DA V2). *: results from Depth Anything V2.}
\label{tab:depth_da_2k}
\begin{center}
\begin{tabular}{lccccccc}
\toprule
Method & Marigold* & 	Geowizard* &  DAT V1*&  ZoeDepth& UniCon-Depth (Ours) & DAT V2 (ViT-L)*
 \\
 \midrule
Accuracy (\%)&	86.8	&88.1&	88.5	&89.1&	90.5&	97.1 \\
\bottomrule
\end{tabular}
\end{center}
\end{table}

\begin{table}[t]
\caption{\textbf{Comparison between synchronous and asynchronous sampling scheduling}. Synchronous: denoising image and depth together while replacing depth input with noisy GT depth at each step. Asynchronous (default): denoising image with clean depth input. We test on \ourmethod{}-Depth for depth-to-image generation.}
\label{tab:sync-async}
\begin{center}

\begin{tabular}{lcccc}
\toprule
Method & FID & 	CLIP Score & AbsRel& $\delta_1$
 \\
 \midrule
Sync&	14.78	&32.45&	17.43&	77.00 \\
Async&	13.21	&32.11&	9.26	&91.02 \\
\bottomrule
\end{tabular}
\end{center}
\end{table}

\begin{figure}[t]
\begin{center}
\includegraphics[width=\linewidth]{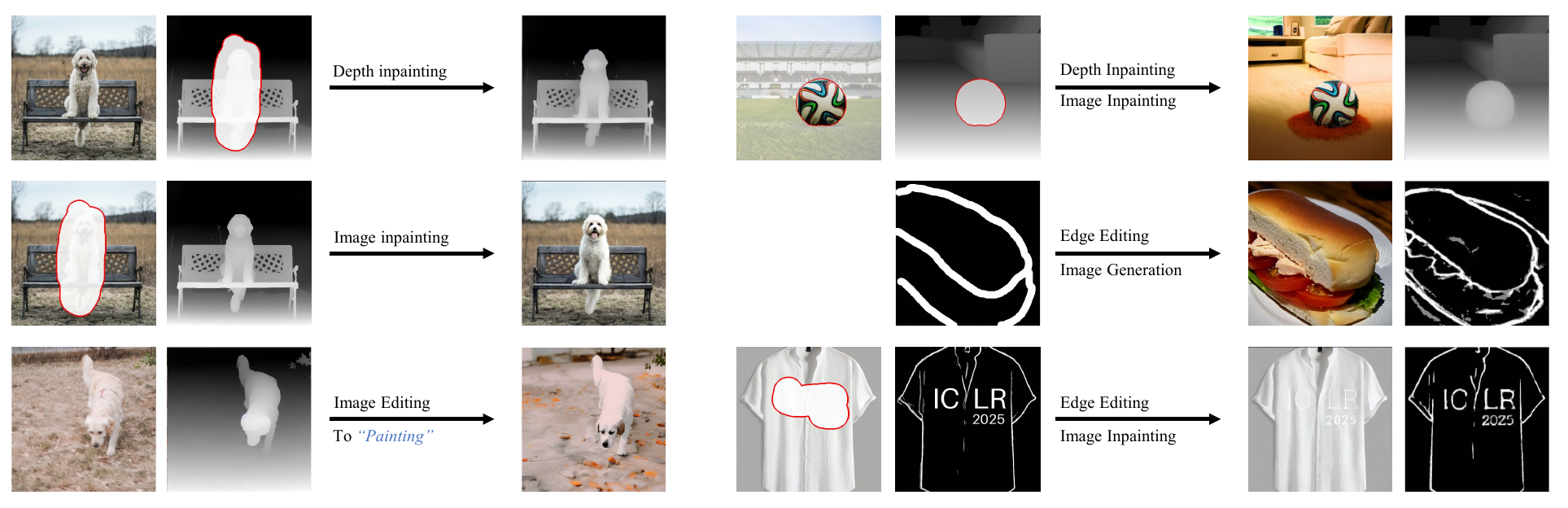}
\end{center}
\caption{\textbf{Inpainting and editing samples.} We show diverse inpainting and editing results using UniCon-Depth and UniCon-Edge.
}
\label{fig:inpainting_editing}
\end{figure}

\subsection{Evaluation Metrics}
For conditional generation (Tab.~\ref{tab:main_res}), we compute Frechet Inception Distance (FID)~\citep{heusel2017gans} between 6K generated images and corresponding real images. We measure condition fidelity using an estimation-matching strategy. In specific, we estimate the condition modalities of the generated images and real images. Then, we compute alignment metrics over the attributes estimated on the generated images and on the real images, to measure the alignment between reference and generated images on the condition modality.

\noindent \textbf{Depth alignment.} We compute the Absolute Mean Relative Error~\citep{ke2023repurposing} (AbsRel) on depth values estimated by Depth-Anything-V2. We adopt the same affine-invariant evaluation protocol as our depth estimation evaluation.

\noindent \textbf{Edge alignment.} For edge alignment, we simply compute the mean squared error on non-zero areas  (\ie edge areas) in the estimated edge maps, thus denoting it as Edge Mean Squared Error (EMSE). Because the edge map is nearly a binary value map, EMSE can directly reflect the alignment of edges.

\noindent \textbf{Pose alignment.} For pose alignment, we compute the standard pose estimation metric Percentage of Correct Keypoints~\citep{yang2012articulated} (PCK) with an adaptation to fit our scenario. PCK measures the alignment between paired ground truth and predicted keypoints. However, the real and generated images in our evaluation may include multiple humans, and we do not have a matching between them. Therefore, we perform a greedy matching between real image keypoints and generated image keypoints. In specific, we compute pair-wise PCK across all sets of keypoints in two images. Then we match the keypoints set greedily in PCK descending order and obtain a matching between two groups of keypoints. Finally, we compute PCK over the matched keypoints.

For depth estimation evaluation, we follow the same affine-invariant evaluation setting as Marigold~\citep{ke2023repurposing}, \ie aligning the prediction with ground truth with the least
squares fitting. Suppose $\rva$ is the predicted depth and $\rvd$ is the GT depth. We compute AbsRel as $\frac{1}{M}\sigma_{i=1}^{M}|\rva_i - \rvd_i| / \rvd_i$ where $M$ is the total number of pixels. Another metric $\delta_1$ is defined as the proportion of pixels satisfying $\max(\rva_i/\rvd_i,\rvd_i/\rva_i)<1.25$.

\subsection{Sampling Setting}
For conditional generation comparison in Tab.~\ref{tab:main_res}, we use the DDIM~\citep{song2020denoising} scheduler with eta=1.0. We sample for 50 steps and use a classifier-free guidance scale of 7.5. We use identical sampling settings for all comparison methods.
For depth estimation in Tab.~\ref{tab:estimation_res}, we use the Euler Ancestral scheduler~\citep{karras2022elucidating} to sample 20 steps. Additionally, we find adding a minor noise (10\% of max noise timestep) to the input image helps improve the estimation quality.

For the qualitative comparison in Fig.~\ref{fig:main_comp}, we use the default sampling setting for specialized methods. JointNet~\citep{zhang2023jointnet} does not support guidance in their official implementation. Therefore, we adopt the same guidance scheme on their model to generate the depth guidance sample. We tune our sampling setting to generate each sample, such as the noise level added to the input depth for the rough depth condition task.

\section{Additional Results}





\begin{table}[t]
\caption{\textbf{Inference Speed Comparison for conditional generation.} We compare \ourmethod{} against ControlNet~\citep{zhang2023adding} on denoising iterations per second. }
\label{tab:eff_c2i}
\begin{center}
\begin{tabular}{lccc}
\toprule
Method & ControlNet & \ourmethod{} (Remove Joint Modules) & \ourmethod{}
 \\
 \midrule
iteration/s	&10.82	&7.65	&5.02
\\
\bottomrule
\end{tabular}
\end{center}
\end{table}

\begin{table}[t]
\caption{\textbf{Latency Comparison for Depth Estimation.} We compare \ourmethod{} against Marigold~\citep{ke2023repurposing}, ZoeDepth~\citep{bhat2023zoedepth}, and Depth Anything V2~\citep{depth_anything_v2} on denoising iterations per second, where Marigold and \ourmethod{} are diffusion-based. The latency is tested on one A800 GPU.}
\label{tab:eff_i2d}
\begin{center}
\begin{tabular}{lcccccc}
\toprule
Method & Marigold	&Marigold(LCM)	&ZoeDepth	&Depth Anything V2	& \ourmethod{}
 \\
 \midrule
latency (A800)&	2.3s	&372ms	&289ms	&91ms&	4.0s
\\
\bottomrule
\end{tabular}
\end{center}
\end{table}

\textbf{User Study.} We conduct a user study on conditional generation performance for depth, softedge, pose, and identity conditions. For each model, users are asked to select their preferred image that aligns with the given image and text condition. We show the result in Fig.~\ref{fig:user-study}, where \ourmethod{} produces comparable results against other specialized methods.

\begin{figure}[t]
\begin{center}
\includegraphics[width=\linewidth]{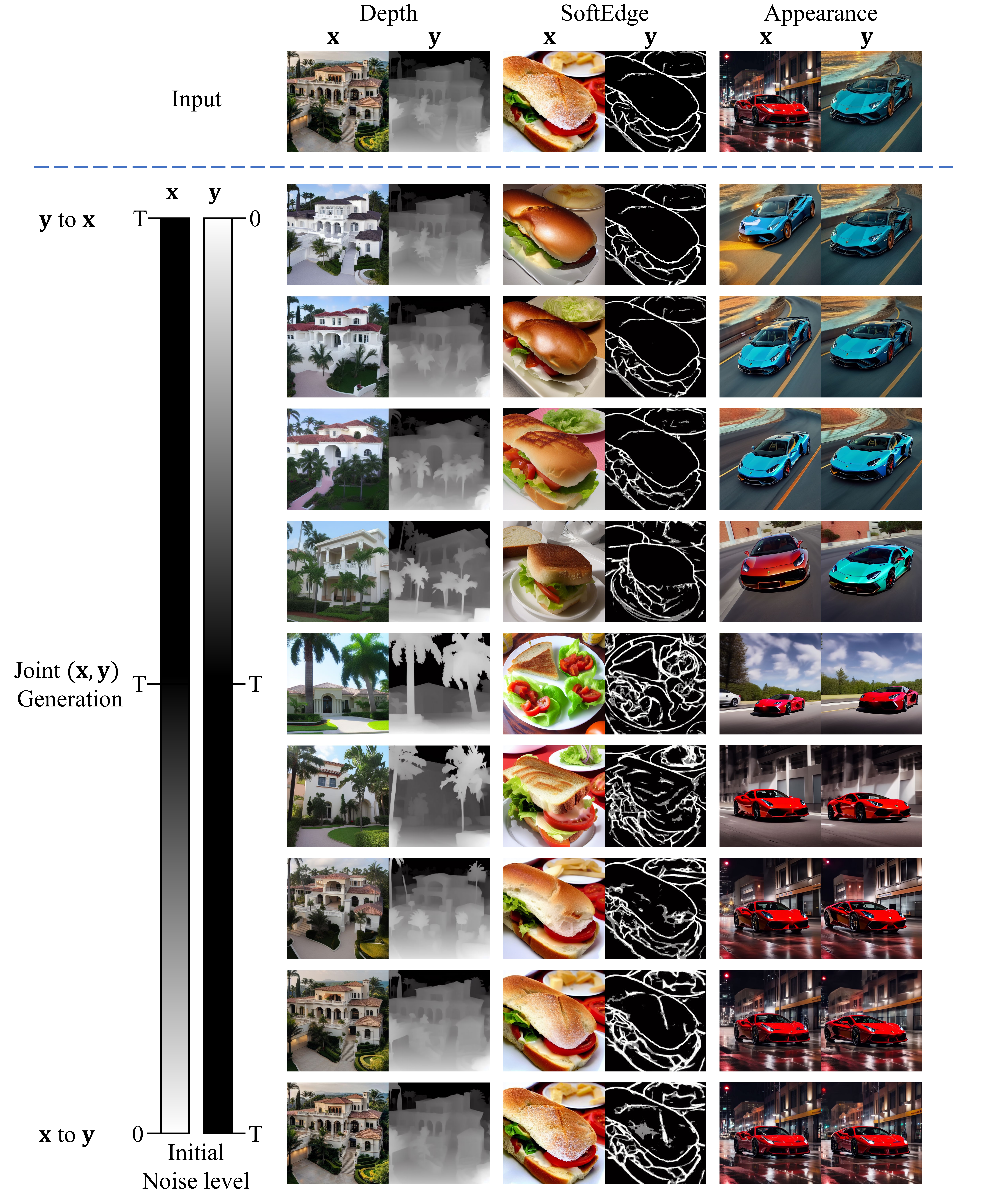}
\end{center}
\caption{\textbf{Interpolating noise schedules.} We can move from $\rvx$-to-$\rvy$ generation to $\rvy$-to-$\rvx$ generation by interpolating the level of initial noise added to $\rvx,\rvy$ inputs. After adding the noise, we denoise $\rvx,\rvy$ together to clean outputs. 
}
\label{fig:interpolate_noise}
\end{figure}

For the Appearance model, evaluating it independently is challenging because (1) Applying it in isolation often produces outputs that closely resemble the input image. We must adjust the condition weight or text prompt for specific use cases. Its effectiveness is most apparent when combined with other models. Please check Fig.~\ref{fig:teaser},\ref{fig:sample_res},\ref{fig:multi-sig},\ref{fig:interpolate_noise} for its use cases.
(2) A suitable baseline for comparison is currently unavailable.

\textbf{Synchronous and asynchronous sampling schedules.} Our disentangled timestep sampling scheme during training enables flexible asynchronous sampling schedules at inference time, which is how \ourmethod{} handles diverse generation tasks. In Tab.~\ref{tab:sync-async}, we compare the depth-to-image generation performance under synchronous and asynchronous sampling schedules.  The results demonstrate that asynchronous noise level scheduling offers clear advantages in depth conditional generation.

For the synchronous setting, we denoise image and depth together while replacing depth input with noisy GT depth at each step. For the asynchronous setting, we denoise the image with clean depth input, which is only possible due to our training strategy.


\noindent \textbf{Quantitative depth estimation.} Though our UniCon model still lags behind state-of-the-art (SOTA) diffusion-based depth estimators (\eg, Marigold~\citep{ke2023repurposing}) on NYUv2 and ScanNet, existing benchmarks often lack scene diversity, as noted in Depth Anything V2~\citep{depth_anything_v2}. To address this limitation, we further evaluate UniCon-Depth on the relative depth estimation benchmark DA-2K, introduced in Depth Anything V2, which comprises diverse high-resolution test images.

On DA-2K (Tab.~\ref{tab:depth_da_2k}), Our UniCon-Depth model is superior to other models except for Depth Anything V2 (which is used as our annotator), demonstrating that our model better generalizes to diverse scenes.

\noindent \textbf{Computation efficiency analysis.} We provide an analysis of our inference cost. Tab.~\ref{tab:eff_c2i} compares UniCon with ControlNet for controllable image generation. UniCon incurs additional inference overhead due to the separate computation of LoRA layers (condition LoRA and joint cross-attention LoRA) instead of fusing them into the pretrained weights. Therefore, separate LoRA layer computation incurs heavy overhead. This overhead can potentially be mitigated by saving multiple attention weights (with and without LoRA) directly in the model.

Tab.~\ref{tab:eff_i2d} compares UniCon with other depth estimation methods. Diffusion-based methods (e.g., UniCon and Marigold) generally exhibit higher latency than traditional approaches. However, UniCon can leverage techniques like Latent Consistency Models (LCM), as demonstrated by Marigold, to significantly improve inference speed.

In summary, while UniCon currently has higher inference costs, its computational efficiency can be enhanced through optimizations like LoRA weight fusion and LCM integration. We plan to explore these directions in future work.

\noindent \textbf{Inpainting and editing using \ourmethod{}.} \ourmethod{} supports diverse editing and inpainting behavior on both image input and condition input, as shown in Fig.~\ref{fig:inpainting_editing}. In addition to inpainting or editing one input conditioning on another, we can simultaneously edit and inpaint both inputs. 

Empirically, we observe that our model is robust for editing tasks. However, its inpainting performance is less stable, which is a limitation of the base model (Stable Diffusion). A straightforward solution is to apply our method to an inpainting-specific diffusion model, such as SD-Inpainting.

\begin{figure}[t]
\begin{center}
\includegraphics[width=\linewidth]{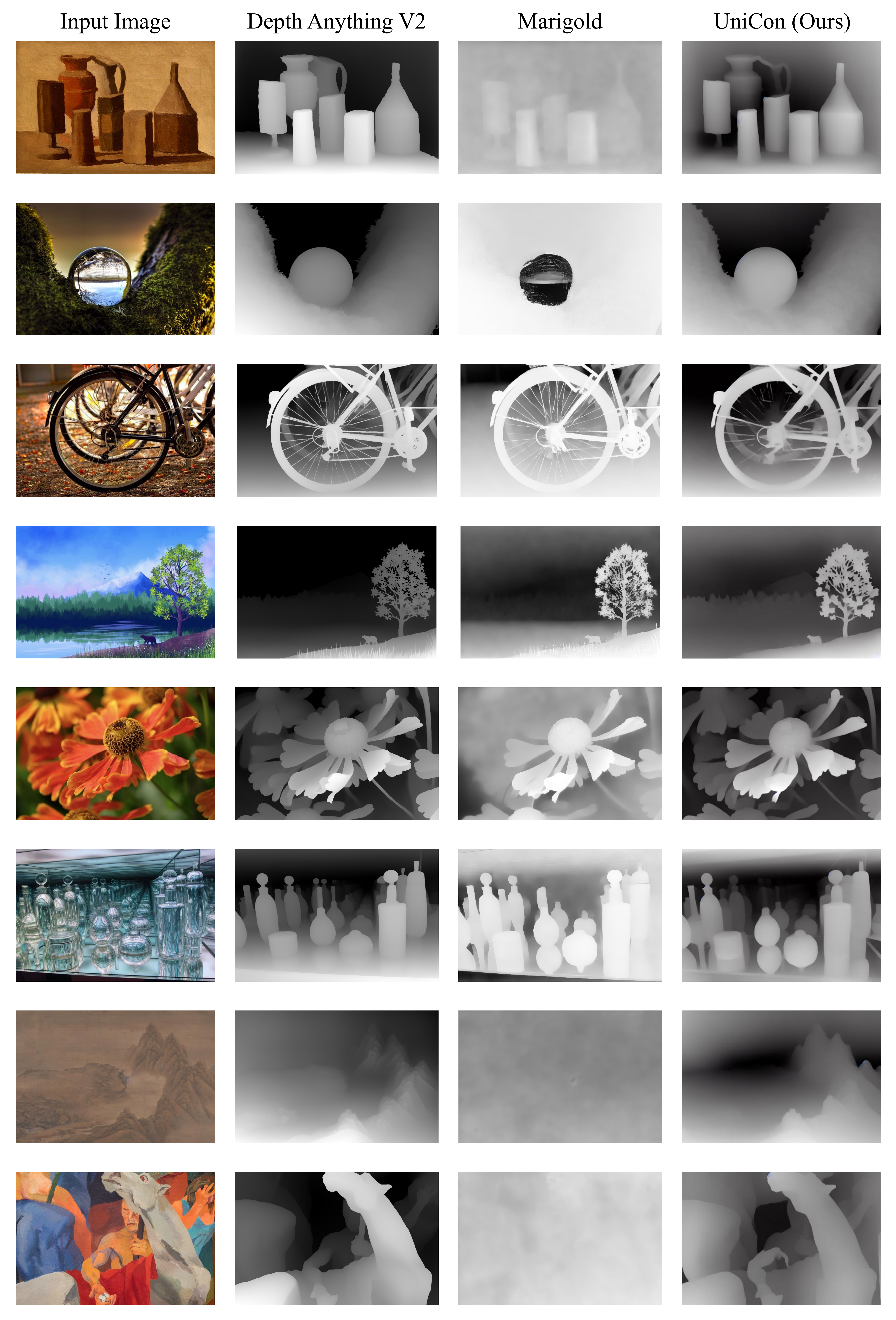}
\end{center}
\caption{\textbf{Qualitative results on depth estimation.} We compare \ourmethod{}-Depth against Depth Anything V2~\citep{depth_anything_v2}, Marigold~\citep{ke2023repurposing}. Test images are from Depth Anything V2~\citep{depth_anything_v2}.
}
\label{fig:qual_i2d}
\end{figure}

\begin{figure}[t]
\begin{center}
\includegraphics[width=\linewidth]{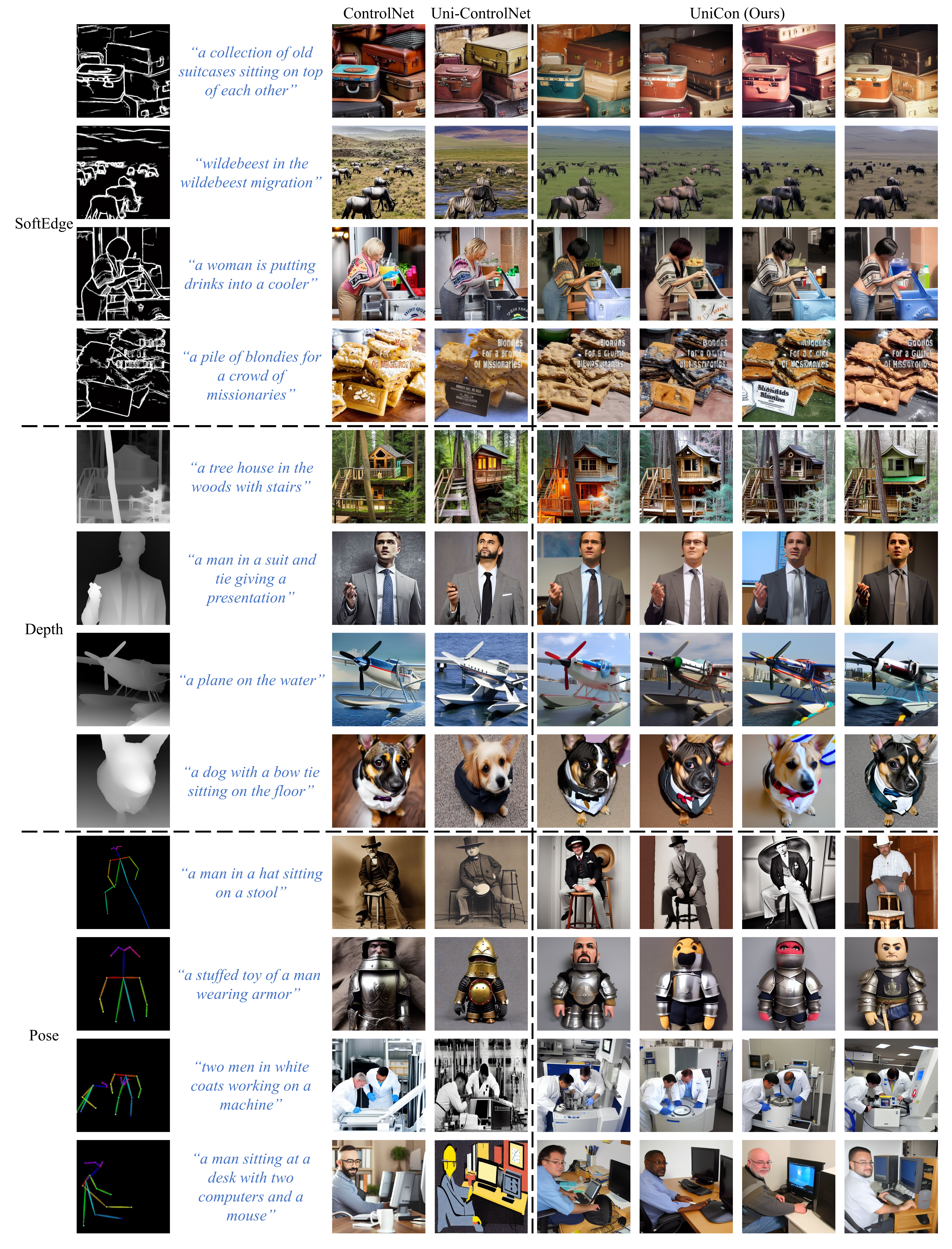}
\end{center}
\caption{\textbf{Qualitative results on conditional image generation.} We compare our conditional image generation results against ControlNet~\citep{zhang2023adding}, Uni-ControlNet~\citep{zhao2024uni}. Uni-ControlNet accepts different inputs for depth and softedge, generated by its default preprocessors (MiDaS, HED).
}
\label{fig:qual_c2i}
\end{figure}

\noindent \textbf{Interpolating between noise schedules.} In Fig.~\ref{fig:interpolate_noise}, we show that we can interpolate the initial noise levels for input $\rvx,\rvy$ to gradually change our sampling behavior in $\rvy$-to-$\rvx$, rough $\rvy$-to-$\rvx$, joint $\rvx,\rvy$ generation, rough $\rvx$-to-$\rvy$, and $\rvx$-to-$\rvy$.

\noindent \textbf{Qualitative results on depth estimation.} We provide more qualitative results on depth estimation for our \ourmethod{}-Depth model (Fig.~\ref{fig:qual_i2d}). Our method performs better than Marigold~\citep{ke2023repurposing} on most test cases, especially for images with ambiguous structure (Fig.~\ref{fig:qual_i2d} Line 2,5) or non-realistic style (Fig.~\ref{fig:qual_i2d} Line 1,7,8). By learning the bidirectional correlation between image and depth, our model better preserves the natural image knowledge of the base model, while fine-tuning the whole model in a conditional sampling manner (as Marigold) might destroy the pretrained image prior.

\noindent \textbf{Qualitative results on conditional image generation.} We provide more qualitative results of our Depth, Edge, and Pose model (Fig.~\ref{fig:qual_c2i}). ControlNet~\citep{zhang2023adding} sometimes generates over-saturated images (SoftEdge 2,4 in Fig.~\ref{fig:qual_c2i}) or images not aligned with the condition (Pose 4 in Fig.~\ref{fig:qual_c2i}), while it rarely happens to our method. We attribute it to that joint modeling stimulates our model to fully capture the bidirectional correlation between image and condition, while conditional modeling might learn to rely on minor clues in the condition image.

\end{document}